\documentclass[11pt,a4paper]{article}
\usepackage[hyperref]{emnlp2018}
\usepackage{times}
\usepackage{latexsym}

\usepackage{amsmath}
\usepackage{url}

\usepackage[T1]{fontenc}
\usepackage{color,soul}
\usepackage{colortbl}

\usepackage{todonotes}
\definecolor{bvd-green}{HTML}{B7FEA6}
\definecolor{bvd-blue}{HTML}{A6D1FE}
\definecolor{bvd-gray}{HTML}{EEE9FD}

\usepackage{multirow}
\usepackage{diagbox}
\usepackage{booktabs}
\usepackage{amssymb}
\usepackage{pifont}
\usepackage{xspace}
\newcommand{\cmark}{\ding{51}\xspace}%
\newcommand{\xmark}{\ding{55}\xspace}%

\newcommand{\semNum}{7\xspace} 
\newcommand{\dataNum}{13\xspace} 
\newcommand{\exampleNum}{over half a million\xspace}
\newcommand{\website}{\url{http://www.decomp.net}}
\newcommand{\inferSent}{\texttt{InferSent}\xspace}

\usepackage{linguex}

\usepackage{subfig}
\usepackage{caption}
\usepackage{cleveref}

\usepackage{diagbox}
\usepackage{adjustbox}
\newcolumntype{H}{>{\setbox0=\hbox\bgroup}c<{\egroup}@{}}

\aclfinalcopy 
\def\aclpaperid{1494} 

\newcommand{\tabref}[1]{Table~\ref{#1}}
\newcommand{\appref}[1]{Appendix~\ref{#1}}

\usepackage{cleveref}

\crefformat{section}{\S#2#1#3}
\newcommand{\secref}[1]{\cref{#1}}

\newcommand{\verbose}[1]{#1}
\renewcommand{\verbose}[1]{}
\newcommand{\cameraready}[1]{#1}
\renewcommand{\cameraready}[1]{#1}

\AtBeginDocument{\setlength{\Exlabelsep}{.1em}}
\AtBeginDocument{\setlength{\Extopsep}{.3\baselineskip}}
\AtBeginDocument{\setlength{\SubExleftmargin}{1.2em}}

\usepackage{setspace}
\usepackage{lipsum}
\usepackage{etoolbox}
\AtBeginEnvironment{quote}{\singlespace\vspace{-\topsep}\small}
\AtEndEnvironment{quote}{\vspace{-\topsep}\endsinglespace}

\title{Collecting Diverse Natural Language Inference Problems\\for  Sentence Representation Evaluation}

\author{Adam Poliak$^{1}$ \hspace{.25cm} Aparajita Haldar$^{1,2}$ \hspace{.25cm} Rachel Rudinger$^{1}$ \hspace{.25cm} J. Edward Hu$^{1}$\\\textbf{Ellie Pavlick$^{3}$  \hspace{.25cm} Aaron Steven White$^{4}$ \hspace{.25cm}Benjamin Van Durme$^{1}$}\\
$^{1}$Johns Hopkins University, $^{2}$ BITS Pilani, Goa Campus, India\\
$^{3}$Brown University, $^{4}$University of Rochester\\ 
}

\date{}

\begin{document}
\maketitle

\begin{abstract}
We present a large scale collection of diverse natural language inference (NLI) datasets
that help provide insight into
how well a
sentence representation captures 
distinct types of reasoning. 
The collection results from recasting 
\dataNum existing datasets 
from \semNum 
semantic phenomena
into a common NLI structure, 
resulting in \exampleNum labeled context-hypothesis pairs in total.
We refer to our collection as the DNC: \textit{\textbf{D}}iverse \textit{\textbf{N}}atural Language Inference \textit{\textbf{C}}ollection.
The DNC is available online at \website, 
and will grow over time as additional resources are recast 
and added from novel sources.
\end{abstract}

\section{Introduction}
A plethora of new natural language inference (NLI)\footnote{The task of determining if a  
hypothesis would likely be inferred from a context, or premise; also known as Recognizing Textual Entailment (RTE)~\cite{dagan2006pascal,dagan2013recognizing}.} datasets has
been created in recent years~\cite{snli:emnlp2015,williams2017broad,lai-bisk-hockenmaier:2017:I17-1,scitail}. 
However, these datasets do not provide clear insight into what type of reasoning or inference a model may be performing. 
For example, these datasets cannot be used to evaluate whether competitive NLI models can determine if an event occurred, correctly differentiate between figurative and literal language, or accurately identify and categorize named entities.
Consequently, these datasets cannot answer how well sentence representation learning models capture distinct semantic phenomena necessary for general natural language understanding (NLU).

To answer these questions, we introduce
the \textbf{D}iverse \textbf{N}LI \textbf{C}ollection (DNC), a large-scale NLI dataset that tests a model's ability to perform diverse types of reasoning. 
The DNC is 
a collection of 
NLI problems, each requiring a model to perform a unique type of reasoning. Each NLI dataset contains labeled context-hypothesis pairs that we  
recast from 
semantic annotations 
for
specific structured prediction tasks.
We extend various prior works on challenge NLI datasets~\cite{zhang2017ordinal}, 
and 
define recasting as leveraging existing datasets to create NLI examples~\cite{glickman2006applied,white-EtAl:2017:I17-1}.
We recast
annotations from a total of \dataNum datasets across \semNum NLP tasks into labeled NLI examples. The tasks include
event factuality, 
named entity recognition, datasets,
gendered anaphora resolution, sentiment analysis, relationship extraction, 
pun detection, and lexicosyntactic inference.
Currently,
DNC contains \exampleNum
labeled examples. 
\tabref{tab:nli_examples} includes 
NLI pairs that test  
specific types of reasoning. 
\begin{table}[t!]
 \centering
 \small
 \renewcommand{\arraystretch}{1.25}
 \begin{tabular}{p{.95cm}p{5.45cm}|p{.01cm}}
     \toprule 
& $\blacktriangleright$ 
Find him before he finds the dog food \\ Event &  \hspace{1em}The finding did not happen & 
\multirow{-2}{*}{\cmark} \\ 
Factuality & $\blacktriangleright$ 
 I'll need to ponder &  \\ 
& \hspace{1em}The pondering happened & \multirow{-2}{*}{\xmark}  \\ \midrule 
& $\blacktriangleright$ Ward 
joined Tom in their native Perth & \\
Relation & \hspace{1em}Ward was born in Perth & \multirow{-2}{*}{\cmark} \\ 
Extraction & $\blacktriangleright$ Stefan had visited his son in Bulgaria & \\ 
& \hspace{1em}Stefan was born in Bulgaria & \multirow{-2}{*}{\xmark} \\ \midrule 
& $\blacktriangleright$ Kim heard masks have no face value & \\ 
& \hspace{1em}Kim heard a pun & \multirow{-2}{*}{\cmark} \\ 
& $\blacktriangleright$ Tod heard that thrift is better than annuity & \\ 
\multirow{-4}{*}{Puns} & \hspace{1em}Tod heard a pun & \multirow{-2}{*}{\xmark} \\ 
\bottomrule 
 \end{tabular}
 \caption{Example sentence pairs for different semantic phenomena. 
$\blacktriangleright$ indicates the line is a context and the following line is its corresponding hypothesis.
\cmark and \xmark respectively indicate that the context entails, or does not entail the hypothesis.
\appref{app:recast-examples} includes 
more recast examples.} 
 \label{tab:nli_examples}
 \end{table}

Using a \emph{hypothesis-only} NLI model, with access to just hypothesis sentences, as a strong baseline~\cite{1804.08117,Gururangan:2018,hypoths-nli}, our experiments demonstrate how 
DNC can be used 
to probe a model's ability to capture 
different types of semantic reasoning necessary for general 
NLU. 
\verbose{We use a \emph{hypothesis-only} NLI model, with access to just hypothesis sentences, as a strong baseline~\cite{1804.08117,Gururangan:2018,hypoths-nli}.}
 In short, this work 
 answers a recent plea to the community to test ``more kinds of inference'' than in previous 
challenge sets~\cite{W17-7203}. 

\verbose{The rest of the paper is organized as follows. In~\secref{sec:motivation}, we motivate recasting NLI and situate it in the larger NLI context. We then describe our efforts to recast \dataNum datasets into NLI in~\secref{sec:recast}. We describe our experimental setup and results in~\secref{sec:experiments} and conclude by describing related work in~\secref{sec:related-word}.}

\section{Motivation \& Background}
\label{sec:motivation}
Compared to 
eliciting NLI datasets directly, i.e. asking humans to author contexts and/or hypothesis sentences, 
recasting
1) help 
determine whether an NLU model
performs distinct types of reasoning\verbose{, 
since converting a prior annotation of a specific phenomenon into an NLI hypothesis results in a focused inference task};
2) limit types of biases observed in previous NLI data; and
3) generate examples cheaply, 
potentially at large scales.

\noindent
\paragraph{NLU Insights} 
Popular NLI datasets, e.g. Stanford Natural Language Inference (SNLI)~\cite{snli:emnlp2015}
and its successor Multi-NLI~\cite{williams2017broad}, were created by eliciting hypotheses from humans. 
Crowd-source workers were tasked with writing one sentence each that is entailed,  
neutral, and contradicted by a 
caption extracted from the Flickr30k corpus~\cite{young2014image}.
\verbose{\footnote{The human-elicitation protocol causes inconsistencies in how different NLI examples deal with some linguistic phenomena, e.g. questions and first/second person. \appref{sec:ling-issues} further describes this issue.}}
Although these datasets are widely used to train and evaluate sentence representations, a high accuracy is not indicative of what types of reasoning NLI models perform. Workers were free to create any type of hypothesis 
for each context and label.
Such datasets cannot be used to determine how well an NLI model captures many desired capabilities of language understanding systems, e.g. paraphrastic inference, complex anaphora resolution~\cite{white-EtAl:2017:I17-1}, 
or compositionality~\cite{P16-1204,2018arXiv180204302D}.
By converting prior annotation of a specific phenomenon into NLI examples, 
recasting allows us to create 
a diverse NLI 
benchmark 
that tests a model's ability to perform distinct types of reasoning. 

\noindent
\paragraph{Limit Biases}
Studies  
indicate that many NLI datasets contain significant biases.
Examples in the 
early Pascal RTE datasets could be correctly predicted
based on syntax alone~\cite{vanderwende2006syntax,vanderwende2006microsoft}. 
Statistical irregularities, and annotation artifacts, within class labels allow a hypothesis-only model 
to significantly outperform the majority baseline 
on at least six recent NLI datasets~\cite{hypoths-nli}. 
Class label biases may 
be attributed to 
 the human-elicited protocol.
Moreover, examples in such NLI datasets 
may contain racial and gendered 
stereotypes~\cite{rudinger-may-vandurme:2017:EthNLP}. 

We limit some biases by not relying on humans to generate hypotheses. Recast NLI datasets may still contain some 
biases, e.g. non-uniform distributions over NLI labels caused by the distribution of labels in the original dataset that we recast.\footnote{
In a corpus 
with part-of-speech tags, the distribution of labels for the word ``the'' will likely peak at the \texttt{Det} tag.} 
Experimental results using \newcite{hypoths-nli}'s hypothesis-only model indicate 
to what degree the recast datasets 
retain 
some biases that may be present in the original semantic datasets. 

\noindent
\paragraph{NLI Examples at Large-scale} 
Generating NLI datasets from scratch is costly. 
Humans must be paid to generate or label natural language text. 
This linearly scales costs
as the amount of generated NLI-pairs 
increases. 
Existing annotations for a wide array of semantic NLP tasks are freely available. By leveraging  
existing semantic annotations 
already invested in by the community 
we can 
generate and label NLI pairs at little cost 
and create large NLI datasets to train data hungry models. 

\noindent
\paragraph{Why These Semantic Phenomena?}
A long term goal is to develop NLU systems that can achieve human levels of understanding and reasoning. Investigating how different architectures and training corpora can help a system perform human-level general NLU is an important step in this direction. 
DNC contains recast NLI pairs 
that are easily understandable by humans and can be used to evaluate different sentence encoders and NLU systems.
These semantic phenomena cover distinct types of reasoning that an NLU system may often encounter in the wild. 
While higher performance on these benchmarks might not be conclusive proof of a system achieving human-level reasoning, 
a system that does poorly
should not be viewed as performing human-level NLU. 
We argue that these semantic phenomena 
play integral roles in NLU. 
There exist 
more semantic phenomena
integral to NLU~\cite{allen1995natural} and we plan to include them in future versions of the DNC.

\noindent
\paragraph{Previous Recast NLI}
Example 
sentences in RTE1~\cite{dagan2006pascal} were extracted from MT, IE, and QA datasets, with the process referred to as  `recasting' in the thesis by~\newcite{glickman2006applied}. 
NLU problems were reframed 
under the NLI framework and 
candidate sentence pairs were extracted from existing NLP datasets and then labeled 
under NLI~\cite{dagan2006pascal}. 
Years later, this term was independently used by \newcite{white-EtAl:2017:I17-1}, who proposed to 
``leverage existing large-scale semantic annotation collections as a source of targeted textual inference examples.'' The term `recasting' was limited to automatically converting existing semantic annotations into labeled NLI examples without manual intervention. 
We 
adopt the broader definition of `recasting' since 
our NLI examples were automatically or manually generated from prior NLU datasets.

\noindent
\paragraph{Applied Framework versus Inference Probing}
Traditionally, NLI has not been viewed as a downstream, applied NLP task.\footnote{This changed as large NLI datasets 
have recently been used to train, or pre-train, models to perform NLI, or other tasks~\cite{conneau-EtAl:2017:EMNLP2017,pasunuru2017multi}.}
Instead, the community has often used it as ``a generic evaluation framework'' to compare models  
for distinct downstream tasks~\cite{dagan2006pascal} or to determine whether a model performs distinct types of reasoning~\cite{cooper1996using}. 
These two 
different evaluation goals may affect which datasets are recast. We 
target both goals as we recast applied tasks 
and 
linguistically focused phenomena. 

\section{Recasting Semantic Phenomena}
\label{sec:recasting}
We describe efforts to recast \semNum semantic phenomena from a total of \dataNum datasets 
 into labeled NLI examples.  
Many of the recasting methods rely on simple templates that do not include nuances and variances typical of natural language. 
This allows us to specifically test how 
sentence representations capture
distinct types of reasoning. 
When recasting, we preserve each dataset's train/dev/test split. If a dataset does not contain such a split, we create 
a random split with roughly a $80$:$10$:$10$ ratio. 
\tabref{tab:NLI-data-stats} reports 
statistics about each 
recast
dataset.

\noindent
\paragraph{Event Factuality (EF)}
Event factuality prediction is the task of determining whether an 
event described in text 
occurred. Determining whether an event occurred
enables accurate inferences, e.g. monotonic inferences,
based on the event~\cite{neural-models-of-factuality}.\footnote{\appref{app:ef} provides an example.}
Incorporating factuality has been shown to improve NLI~\cite{saurf2007determining}.

We recast event factuality annotations from
UW~\cite{lee2015event}, MEANTIME~\cite{minardmeantime}, and 
Decomp~\cite{neural-models-of-factuality}. 
We use sentences from the original datasets as
contexts and 
templates \ref{ex:ef-hyp} and \ref{ex:ef-not-hyp} as hypotheses.\footnote{We replace \textit{Event} with the event described in the context.} 
\ex. 
\a. The \textit{Event} happened
\label{ex:ef-hyp}
\b. The \textit{Event} did not happen
\label{ex:ef-not-hyp}

If the predicate denoting the \textit{Event} was  
annotated as having happened  
in the 
factuality dataset, the context paired with \ref{ex:ef-hyp} is labeled as \textsc{entailed} and the same context paired with \ref{ex:ef-not-hyp} is labeled as \textsc{not-entailed}.
Otherwise, we swap the labels. 

\begin{table*}[t!]
\centering
\small
\begin{tabular}{c|cH|c|c}
\hline
Sem. Phenomena  
& Dataset & \# Contexts & \# pairs & Automated\\ \toprule 
& Decomp~\cite{neural-models-of-factuality} 
 & & 42K (41,888) & \cmark \\ 
& UW~\cite{lee2015event} & & 5K (5,094) & \cmark \\  
\multirow{-3}{*}{Event Factuality} & MeanTime~\cite{minardmeantime} & & .7K (738) & \cmark \\ \midrule 
& Groningen~\cite{bos2017groningen} & 130K (130,703) & 260K (261,406) & \cmark \\ 
\multirow{-2}{*}{Named Entity Recognition} & CoNLL~\cite{TjongKimSang:2003:ICS:1119176.1119195} & 30K (29985) & 60K (59,970) & \cmark \\ \midrule 
Gendered Anaphora & Winogender~\cite{gender-bias-in-coreference-resolution} & & .4K (464) & \xmark \\ \midrule
 & 
 VerbCorner~\cite{hartshorne2013verbcorner} & &  $135$K ($138,648$) & \cmark\\ 
&  MegaVeridicality~\cite{white_role_2018} & & 11K (11,814) & \cmark \\
\multirow{-3}{*}{Lexicosyntactic Inference}  & VerbNet~\cite{schuler2005verbnet} & & 2K ($1,759$) & \cmark \xmark\\ 
\midrule 
& ~\cite{D15-1284} & & 9K (9,492) & \cmark\\
\multirow{-2}{*}{Puns} & SemEval 2017 Task 7~\cite{miller-hempelmann-gurevych:2017:SemEval}  & & $8$K ($8,054$) & \cmark \\ \midrule 
Relationship Extraction & FACC1~\cite{gabrilovich2013facc1}& & $25$K (25,132) & \cmark \xmark \\ \midrule 
Sentiment Analysis & \cite{kotzias2015group} & & 6K (6,000) & \cmark \\ \midrule
\midrule
Combined & Diverse NLI Collection (DNC) & & 570K (570,459)\\ \midrule  
\midrule 
--- & SNLI~\cite{snli:emnlp2015}  & & 570K \\
--- & Multi-NLI~\cite{williams2017broad} & & 433K\\ \bottomrule 
\end{tabular}
\caption{Statistics summarizing the recast datasets. The first column refers to the original annotation that was recast, the `Combined` row refers to the combination of our recast datasets. The second column indicates the datasets that were recast, and the 3rd column reports how many labeled NLI pairs were extracted from the corresponding dataset. 
The last column indicates whether the recasting method was fully-automatic without human involvement (\cmark), manual (\xmark), or used a semi-automatic method that included human intervention (\cmark \xmark). 
The Multi-NLI and SNLI numbers contextualize the scale of our dataset.}
\label{tab:NLI-data-stats}
\end{table*}

\noindent
\paragraph{Named Entity Recognition (NER)}
Distinct types of entities have different properties and relational objects~\cite{prince1978function}\verbose{. 
Relational objects} that can  
help infer facts from a given context. 
For example, if a system can detect that an entity is a name of a nation, then that entity likely has
a leader, a language, and a culture~\cite{prince1978function,extracting-implicit-knowledge-from-text}. 
When classifying NLI pairs, a model can determine if an object mentioned in the hypothesis can be a relational object typically associated with the type of entity described in the context. 
NER tags can also be directly used to determine if a hypothesis is likely to not be entailed by a context, such as when entities in contexts and hypotheses do not share NER tags \verbose{, e.g. \texttt{PERSON} or \texttt{ORGANIZATION}.
This intuition has been successfully used in NLI}~\cite{castillo2008approach,sammons2009relation,pakray2010ju_cse_tac}. 

Given a sentence annotated with NER tags, 
we recast the annotations by
preserving the original sentences as contexts and creating hypotheses using the template 
``\textit{NP} is a \textit{Label}.''\footnote{We ensure grammatical hypotheses by appropriately conjugating ``is a'' when needed.}
For 
\textsc{entailed} hypotheses we replace \textit{Label} with the correct NER label of the \textit{NP}; 
for \textsc{not-entailed} hypotheses, we choose an incorrect label from 
the prior distribution of NER tags for the given phrase.
This prevents us from adding additional 
biases besides any class-label statistical irregularities present in the original 
data.
We apply this  
procedure on the
Gronigen Meaning Bank~\cite{bos2017groningen} and the ConLL-2003  
Shared Task~\cite{TjongKimSang:2003:ICS:1119176.1119195}.

\noindent
\paragraph{Gendered Anaphora Resolution (GAR)}
The ability to perform pronoun resolution is essential to language understanding, in many cases requiring common-sense reasoning about the world~\citep{levesque2012winograd}. 
\citet{white-EtAl:2017:I17-1} show that this task can be directly recast as an NLI problem by transforming Winograd schemas into NLI sentence pairs.

Using a similar formula 
\citet{gender-bias-in-coreference-resolution} introduce \textit{Winogender schemas}, minimal sentence pairs that differ only by pronoun gender. With this adapted pronoun resolution task, they demonstrate the presence of systematic gender bias in coreference resolution systems. 
We recast Winogender schemas as an NLI task, introducing a potential method of detecting gender bias in NLI systems or sentence embeddings. In recasting, the context is the original, unmodified Winogender sentence; the hypothesis is a short, manually constructed sentence having a
correct (\textsc{entailed}) or incorrect (\textsc{not-entailed}) pronoun resolution.

\noindent
\paragraph{Lexicosyntactic Inference (Lex)}
While many inferences in natural language are triggered by lexical items alone, there exist pervasive inferences that arise from interactions between lexical items and their syntactic contexts. This is particularly apparent among propositional attitude verbs -- e.g. \textit{think}, \textit{want}, \textit{know} -- which display complex distributional profiles~\citep{white2016computational}. For instance, the verb \textit{remember} can take both finite clausal complements 
and infinitival clausal complements.

\ex.
\a. Jo didn't \textbf{remember} \textit{that she ate}
\label{ex:rememberthat}
\b. Jo didn't \textbf{remember} \textit{to eat}
\label{ex:rememberto}

This small change in the syntactic structure gives rise to large changes in the inferences that are licensed: \ref{ex:rememberthat} presupposes that \textit{Jo ate} while \ref{ex:rememberto} entails that \textit{Jo didn't eat}. 
We recast data 
from three datasets that are relevant to these sorts of lexicosyntactic interactions. 

\noindent
\textbf{Lex \#1: MegaVeridicality (MV)} \citet{white_role_2018} build the MegaVeridicality dataset by selecting verbs from the MegaAttitude dataset \citep{white2016computational} based on their grammatical acceptability in the [NP \_ that S] and [NP was \_ed that S] frames.\footnote{NP is always instantiated by \textit{someone}; and S is always instantiated by \textit{a particular thing happened}.} They then asked annotators to answer questions of the form in \ref{ex:lexicosyntactictask} using three possible responses: \textit{yes}, \textit{maybe or maybe not}, and \textit{no} \citep{karttunen_chameleon-like_2014}. 

\ex. \label{ex:lexicosyntactictask}
\a. Someone \{knew, didn't know\} that a particular thing happened. \label{ex:lexicosyntactictaskcontext}
\b. Did that thing happen?

We use the same procedure to annotate sentences containing verbs that take various types of infinitival complement: [NP \_ for NP to VP], [NP \_ to VP], [NP \_ NP to VP], and [NP was \_ed to VP].\footnote{NP is always instantiated by either \textit{someone}, \textit{a particular person}, or \textit{a particular thing}; and VP is always instantiated by \textit{happen}, \textit{do a particular thing}, or \textit{have a particular thing}.}

To recast these annotations, we assign the context sentences like \ref{ex:lexicosyntactictaskcontext} to the majority class -- \textit{yes}, \textit{maybe or maybe not}, \textit{no} -- across 10 different annotators, after applying an ordinal model-based normalization to their responses. We then pair each context sentence with three hypotheses. 

\ex.
\a. That thing happened \label{ex:lexicosyntacticyes}
\b. That thing may or may not have happened \label{ex:lexicosyntacticmaybe}
\c. That thing didn't happen \label{ex:lexicosyntacticno}

If annotated \textit{yes}, \textit{maybe or maybe not}, or \textit{no}, the pair \ref{ex:lexicosyntactictaskcontext}-\ref{ex:lexicosyntacticyes}, \ref{ex:lexicosyntactictaskcontext}-\ref{ex:lexicosyntacticmaybe}, or \ref{ex:lexicosyntactictaskcontext}-\ref{ex:lexicosyntacticno} is respectively assigned \textsc{entailed} and the other pairings are assigned \textsc{not-entailed};
train/dev/test split labels are randomly assigned 
to every pair that context sentence appears in. 

\paragraph{Lex \#2: Recasting VerbNet (VN)}
We create additional lexicosyntactic NLI examples from VerbNet~\cite{schuler2005verbnet}. VerbNet contains classes of verbs that each can have multiple frames.
Each frame contains a mapping from syntactic arguments to thematic roles, 
which are used as arguments in 
Neo-Davidsonian first-order logical predicates~\ref{vn:pred-main} 
that describe the frame's semantics. Each frame additionally contains an example sentence~\ref{vn:context-main} that we use as our NLI context and we create templates~\ref{vn:template-main} from the most frequent semantic predicates to generate hypotheses~\ref{vn:hyp-main}. 

\ex.
\a. Michael swatted the fly
\label{vn:context-main}
\b. cause(\textit{E}, \textit{Agent})
\label{vn:pred-main}
\c. \textit{Agent} caused the \textit{E}
\label{vn:template-main}
\d. Michael caused the swatting
\label{vn:hyp-main}

We use the Berkeley Parser~\cite{petrov2006learning} to match tokens in an 
example sentence with the thematic roles 
and then fill in the templates with the matched tokens~\ref{vn:hyp-main}. We also decompose multi-argument predicates into unary predicates to increase the number of hypotheses we generate. On average, each context is paired with $4.5$ hypotheses. 
We generate \textsc{not-entailed} hypotheses by filling in templates with incorrect thematic roles.  
\footnote{
This is similar to \newcite{aharon2010generating}'s
template matching 
to generate 
entailment rules from FrameNet~\cite{baker1998berkeley}.}
We partition the recast NLI examples into train/development/test splits such that all example sentences from a VerbNet class (which we use a NLI hypothesis) appear in only one partition of our dataset. In turn, the recast VerbNet dataset's partition is not exactly 80:10:10. 

\paragraph{Lex \#3: Recasting VerbCorner (VC)} The third dataset testing lexicosyntactic inference that we recast is VerbCorner (VC)~\cite{hartshorne2013verbcorner}. VC decomposes VerbNet predicates into simple semantic properties and 
``elicit[s] reliable semantic judgments corresponding to
VerbNet predicates''
via crowd-sourcing. The semantic judgments focus on
movement, physical contact, application of force, change of physical or mental state, and valence, 
all of which ``may be central organizing principles for a human's 
$\ldots$ conceptualization of the world.''~\cite{hartshorne2013verbcorner}.

Each sentence in VC is judged based on the decomposed semantic properties.
We convert each semantic property into declarative statements\footnote{We list the declarative statements in~\appref{app:vc}.} to create hypotheses 
and pair them
with the original sentences which we preserve as contexts. The NLI pair is \textsc{entailed} or \textsc{not-entailed} depending on the given sentence's semantic judgment.

\noindent
\paragraph{Figurative Language (Puns)} 
Figurative language demonstrates natural language's expressiveness and wide variations. 
Understanding and recognizing figurative language 
``entail[s] cognitive capabilities to abstract and meta-represent meanings beyond \textit{physical} words''~\cite{reyes2012humor}. 

Puns are prime examples of figurative language that may
perplex general NLU systems 
as they are one of the more regular uses of linguistic ambiguity~\cite{binsted1996machine} 
and rely on a wide-range of phonetic, 
morphological, syntactic, and semantic ambiguity~\cite{pepicello1984language,binsted1996machine,bekinschtein2011clowns}. 
 
We recast puns from \newcite{D15-1284} and \newcite{miller-hempelmann-gurevych:2017:SemEval}
using templates to generate contexts~\ref{ex:pun-context} and hypotheses~\ref{ex:pun-hyp}, \ref{ex:pun-not-hyp}. 
We replace \textit{Name} with names sampled from a distribution based on 
US census data,\footnote{\url{http://www.ssa.gov/oact/babynames/names.zip}} 
and 
\textit{Pun} with the original sentence. 
If the original sentence was labeled as containing a pun, the \ref{ex:pun-context}-\ref{ex:pun-hyp} pair
is labeled as \textsc{entailed} and 
\ref{ex:pun-context}-\ref{ex:pun-not-hyp} 
is labeled as
\textsc{not-entailed}, otherwise
we swap the labels. 
\verbose{In total, we generate roughly $15$K labeled pairs.}
\ex.
\a. \textit{Name} heard that \textit{Pun}
\label{ex:pun-context}
\b. \textit{Name} heard a pun
\label{ex:pun-hyp}
\c. \textit{Name} did not hear a pun
\label{ex:pun-not-hyp}

\noindent
\paragraph{Relation Extraction (RE)}
The goal of the relation extraction (RE) task is to infer the real-world relationships between pairs of entities from natural language text. The task is ``grounded'' in the sense that the input is natural language text and the output is $\langle \texttt{entity1}, \texttt{relation}, \texttt{entity2} \rangle$ tuples defined in the schema of some knowledge base. RE requires a system to understand the many different surface forms which may entail the same underlying relation, and to distinguish those from surface forms which involve the same entities but do not entail the relation of interest. For example, \ref{ex:kb-hyp} is entailed by 
\ref{ex:kb-true-ctx} and \ref{ex:kb-true-ctx2} but not by \ref{ex:kb-false-ctx}.

\ex. 
\a. \textit{Name} was born in \textit{Place}
\label{ex:kb-hyp}
\b. \textit{Name} is from \textit{Place}
\label{ex:kb-true-ctx}
\c. \textit{Name}, a \textit{Place} native, $\dots$
\label{ex:kb-true-ctx2}
\d. \textit{Name} visited \textit{Place}
\label{ex:kb-false-ctx}

\noindent Natural language surface forms are often used in RE in a weak-supervision setting \cite{mintz-EtAl:2009:ACLIJCNLP,hoffmann-EtAl:2011:ACL-HLT2011,riedel-EtAl:2013:NAACL-HLT}. 
That is, if \texttt{entity1} and \texttt{entity2} are known to be related by \texttt{relation}, it is assumed that every sentence observed which mentions both \texttt{entity1} and \texttt{entity2} is assumed to be a realization of \texttt{relation}: i.e. \ref{ex:kb-false-ctx} would (falsely) be taken as evidence of the \texttt{birthPlace} relation. 

Here we first generate hypotheses and then corresponding contexts. 
To generate hypotheses, we begin with entity-relation triples extracted from DBPedia infoboxes: e.g. $\langle \textit{Barack Obama}, \texttt{ birthPlace}, \textit{ Hawaii} \rangle$. These relation predicates were extracted directly from Wikipedia infoboxes and are not cleaned. As a result, many relations are redundant with one another (\texttt{birthPlace}, \texttt{hometown}) and some relations do not correspond to obvious natural language glosses based on the name alone (\texttt{demographics1Info}). Thus, we construct a template for each predicate \texttt{p} by manually inspecting 1) a sample of entities which are related by \texttt{p} 2) a sample of sentences in which those entities co-occur and 3) the most frequent natural language strings which join entities related by \texttt{p} according to a OpenIE triple database \cite{schmitz2012open,fader2011identifying} extracted from a large text corpus. 
We then manually write
a simple template (e.g. Mention1 \textit{ was born in } Mention2) for \texttt{p}, ignoring any unclear relations. 
In total, we end up with 574 unique relations, expressed by 354 unique templates.

For each such hypothesis generated, 
we create a number of contexts. 
We begin with the FACC1 corpus \cite{gabrilovich2013facc1} which contains natural language sentences from ClueWeb in which entities have been automatically linked to disambiguated Freebase entities, when possible.
Then, given a tuple $\langle \texttt{entity1}, \texttt{relation}, \texttt{entity2} \rangle$, we find every sentence which contains both \texttt{entity1} and \texttt{entity2}.\verbose{\footnote{This results in a dataset with hypotheses that appear multiple times, similar to \newcite{scitail}'s SciTail.}
\aparajita{footnote applicable to other recast datasets too}}
Since many of these sentences are false positives~\ref{ex:kb-false-ctx},  
we have human annotators vet each context/hypothesis pair, using the ordinal entailment scale described in \newcite{zhang2017ordinal}. 
We include optional binary labels by converting pairs labeled as $1-4$ and $5$ to \textsc{entailed} and \textsc{not-entailed} respectively.\footnote{Following the label set in SNLI, \newcite{zhang2017ordinal} converted pairs labeled with $1$ as \textsc{contradiction}, $2-4$ as \textsc{neutral} and $5$ to \textsc{entailment}. Since here we are generally interested in binary classification, we merge the \textsc{contradiction} and \textsc{neutral} examples as \textsc{not-entailed}.} We apply pruning methods  (described in~\appref{app:kg}) to combat issues related to noisy, ungrammatical hypotheses 
and disagreement between multiple annotators.

\noindent
\paragraph{Subjectivity (Sentiment)}
Some of the previously discussed semantic phenomena 
deal with objective information -- 
did an event occur or what type of entities does a specific name represent.
Subjective information is often expressed differently
~\cite{wiebe2005annotating}, making it important to use other tests to probe whether an NLU system understands language that expresses subjective information.
We are interested in determining whether general NLU models capture `subjective clues' that can help identify and understand 
emotions, opinions, and sentiment within a subjective text~\cite{wilson2006recognizing}\verbose{, as opposed to
differentiating between subjective and objective information~\cite{yu2003towards,riloff2003learning}}.  

We recast a sentiment analysis dataset since the task is the ``expression of subjectivity
as either a positive or negative opinion''~\cite{taboada2016sentiment}. We extract 
sentences from product, movie, and restaurant reviews 
labeled as containing positive or negative sentiment~\cite{kotzias2015group}.
Contexts \ref{ex:sentiment-context} and hypotheses \ref{ex:sentiment-pos-hyp}, \ref{ex:sentiment-neg-hyp} are generated using the following templates: 
\ex. 
\a. When asked about \textit{Item}, \textit{Name} said \textit{Review}
\label{ex:sentiment-context}
\vspace{-12.5pt}
\b. \textit{Name} liked the \textit{Item}
\label{ex:sentiment-pos-hyp}
\c. \textit{Name} did not like the \textit{Item}
\label{ex:sentiment-neg-hyp}

\textit{Item} is replaced with either ``product'', ``movie'', or ``restaurant'', 
and the \textit{Name} is sampled as previously discussed.
If the original sentence 
contained 
positive (negative) sentiment, the \ref{ex:sentiment-context}-\ref{ex:sentiment-pos-hyp} pair is labeled as \textsc{entailed} (\textsc{not-entailed}) and 
\ref{ex:sentiment-context}-\ref{ex:sentiment-neg-hyp} 
is labeled as
\textsc{not-entailed} (\textsc{entailed}). 

\definecolor{light-gray}{gray}{0.97}
\begin{table*}
	\centering
\begin{tabular}{l|cccccc|ccc}
\toprule 
\multicolumn{1}{l|}{\diagbox[width=9em]{Model}{Recast Data}} & NER & EF & RE & Puns & Sentiment & GAR & VC & MV & VN \\
\midrule 
Majority (MAJ) & 50.00 & 50.00 & 59.53 & 50.00 & 50.00 & 50.00 & 50.00 & 66.67 & 53.66 \\
\multicolumn{10}{c}{\cellcolor{light-gray} No Pre-training} \\ 
InferSent & \textbf{92.50} & 83.07 & 61.89 & 60.36 & 50.00 & -- & 88.60 & \textbf{85.96} & 46.00 \\
Hyp-only & 91.48 & 69.14 & 64.78 & 60.36 & 50.00 & -- & 76.82 & 77.83 & 46.00 \\
\multicolumn{10}{c} 
{\cellcolor{light-gray} Pre-trained DNC} \\
InferSent \textit{(update)} & 92.47 & \textbf{83.86} & 74.38 & \textbf{93.17} & 81.00 & -- & \textbf{89.00} & 85.62 & 76.83 \\
InferSent \textit{(fixed)} & 92.20 & 81.07 & 74.11 & 87.76 & 77.33 & \textbf{50.65} & 88.59 & 83.84 & 67.68 \\
Hyp-only \textit{(update)} & 91.60 & 71.07 & 70.57 & 60.02 & 46.83 & -- & 76.78 & 77.83 & 68.90 \\
Hyp-only \textit{(fixed)} & 91.37 & 69.74 & 65.97 & 56.44 & 48.17 & 50.00 & 76.78 & 77.83 & 59.15 \\
\multicolumn{10}{c} {\cellcolor{light-gray} Pre-trained Multi-NLI} \\
InferSent \textit{(update)} & 92.37 & 83.03 & \textbf{76.08} & 92.48 & \textbf{83.50} & -- & 88.45 & 85.11 & \textbf{78.05} \\
InferSent \textit{(fixed)} & 52.99 & 54.88 & 66.75 & 56.04 & 56.50 & \textbf{50.65} & 45.33 & 55.92 & 45.73 \\
Hyp-only \textit{(update)} & 91.62 & 70.64 & 69.91 & 60.36 & 49.33 & -- & 76.82 & 77.83 & 68.29 \\
Hyp-only \textit{(fixed)} & 52.55 & 66.33 & 52.96 & 60.59 & 50.00 & 50.43 & 41.31 & 46.28 & 48.78 \\
\bottomrule
\end{tabular}
\caption{NLI accuracies on test data. 
Columns correspond to each semantic phenomena and rows correspond to the model used. Columns are ordered from larger to smaller in size, but the last three (VC, MV, VN) are separated since they fall under lexico-syntactic inference. \textit{(update)} refers to a model that was initialized with pre-trained parameters and then re-trained on the corresponding recast data. \textit{(fixed)} refers to a model that was trained and then evaluated on these data sets. Bold numbers in each column indicate which settings were responsible for the highest accuracy on the specific recast dataset.} 
\label{tab:nli-results}
\end{table*}

\subsection{Noise in Recast Data}
Recasting can create noisy NLI examples that may potentially enable a model to achieve a high accuracy by learning dataset specific characteristics that are unrelated to NLU.
For example, \newcite{poliakNAACL18,hypoths-nli} previously noted the association between ungrammaticality and \textsc{not-entailed} examples based on how \newcite{white-EtAl:2017:I17-1} recast the FrameNet+ dataset~\cite{pavlick-EtAl:2015:ACL-IJCNLP2}.

In the DNC, most of the noisy examples are in the recast VerbNet and Relation Extraction portions. In recast VerbNet, some examples are noisy because of incorrect subject-verb agreement.\footnote{``Her \textit{teeth was} cared for'' or ``\textit{Floss were} used''.}
Since more noisy examples appeared in the Relation Extraction set, we relied on Amazon Mechanical Turk workers to flag ungrammatical hypotheses in the recast dataset, and we remove NLI pairs with ungrammatical hypotheses.\footnote{See~\appref{app:kg} for details.}

\section{Experiments}
\label{sec:experiments}
Our experiments demonstrate how these recast datasets may be used to evaluate how well models capture different types of semantic reasoning necessary for general language understanding. We also include results from a hypothesis-only model 
as a strong baseline. This 
may 
reveal whether the recast datasets retain statistical irregularities 
from the original, task-specific annotations.

\subsection{Models}
For demonstrating how well an NLI model performs these fine-grained types of reasoning, we use \inferSent~\cite{conneau-EtAl:2017:EMNLP2017}. \inferSent independently encodes a context and hypothesis with a bi-directional LSTM and combines the sentence representations by concatenating the individual sentence representations, their element-wise subtraction and product. The combined representation is then fed into a MLP with a single hidden layer. The hypothesis-only model is a modified version of \inferSent that only accesses hypotheses~\cite{hypoths-nli}. We report experimental details in~\appref{app:sec-exp-details}.
\verbose{This  determines whether a model can perform well on each recast NLI dataset when only considering the context.\footnote{We use the same training settings and hyper-parameters as 
\newcite{conneau-EtAl:2017:EMNLP2017} and \newcite{hypoths-nli}.}}

\subsection{Results}
 \tabref{tab:nli-results} reports the models' accuracies across the recast NLI datasets. 
Even though we categorize 
VerbNet, MegaVeridicality, and VerbCorner as 
lexicosyntatic inference, we train and evaluate models separately on these three datasets because we use different strategies to individually recast them.
When evaluating NLI models, our baseline is the maximum between the 
accuracies of the hypothesis-only model and 
the majority class label (MAJ). 
In six of the eight recast datasets that we use to train our models the hypothesis-only model outperforms MAJ. The two datasets where the hypothesis-only model does not outperform MAJ are Sentiment and VN, each of which contain less than $10K$ examples.\footnote{This is similar to \newcite{hypoths-nli}'s results where a hypothesis-only model did not outperform MAJ on datasets with $\leq$ $10$K examples.}  
We do not train 
on GAR because of its small size. 

Our results suggest that \inferSent, when not pre-trained on any other data, might capture specific semantic phenomena 
better than other semantic phenomena. 
\inferSent seems to learn the most about determining if an event occurred,
since the difference between its accuracy and that of the hypothesis-only baseline (+13.93) is largest on the recast EF
dataset compared to the other recast annotations. The model seems to similarly learn to perform (or detect) the type of lexico-syntactic inference
present in VC and MV. Interestingly,  the hypothesis-only model outperforms \inferSent on the recast 
RE.

\noindent
\paragraph{Hypothesis Only Baseline}
The hypothesis-only model can demonstrate how likely it is that an NLI label applies to a hypothesis, regardless of its context 
and indicates how well each recast dataset tests a model's ability to perform each specific type of reasoning when performing NLI.  
The high hypothesis-only accuracy on the recast NER dataset may demonstrate that the hypothesis-only model is able to detect that the distribution of class labels for a given word may be peaky. For example, \textit{Hong Kong} 
appears $130$ times in the training set and is always labeled as a location. Based on this, in future work we may 
consider different methods to recast NER annotations into labeled NLI examples, or limit the dataset's training size.

\noindent
\paragraph{Pre-training models on DNC}
We would like to know whether initializing models with pre-trained parameters improves scores.\verbose{We pre-train our models on DNC either 
all of DNC or a subset of DNC where we sample at most $10$K (or $20$K) examples from each category}
We notice that when we pre-train our models on DNC, for the larger datasets, a pre-trained model does not seem to significantly outperform
randomly initializing the parameters. 
For the smaller datasets, specifically 
Puns, Sentiment and VN, a pre-trained model significantly outperforms random initialization.\footnote{
By $32.81$, $31.00$, and $30.83$ points respectively.}

We are also interested to know whether fine-tuning these pre-trained models on each category (\textit{update}) improves a model's ability to perform well on the category compared to keeping the pre-trained models' parameters static (\textit{fixed}).
Across all of the recast datasets, updating the pre-trained model's parameters during training improves \inferSent's accuracies more than keeping the model's parameters fixed. 
When updating a model pre-trained on the entire DNC, we see the largest improvements on VN (+9.15).

\noindent
\paragraph{Models trained on Multi-NLI}
\newcite{williams2017broad} argue that Multi-NLI ``[makes] it possible to evaluate
systems on nearly the full complexity of
the language.'' However, how well does Multi-NLI test a model's capability to understand 
the diverse semantic phenomena captured in DNC? 
We posit that if a model, trained on and performing well on Multi-NLI, does not perform well on our recast datasets, then 
Multi-NLI might not evaluate a model's ability to understand the ``full complexity'' of language as argued.\footnote{We treat Multi-NLI's \textsc{neutral} and \textsc{contradiction} labels as equivalent to the DNC's \textsc{not-entailed} label.}

When trained on Multi-NLI, our \inferSent model achieves an accuracy of 70.22\% on (matched) Multi-NLI.\footnote{Although 
this is about $10$ points below SoTA,
we believe that the pre-trained model performs well enough 
to evaluate whether Multi-NLI tests a model's capability to understand the diverse semantic phenomena in the DNC.} 
When we test the model on the recast datasets (without updating the parameters), we see significant drops.\footnote{\inferSent (\textit{pre-trained, fixed}) in ~\tabref{tab:nli-results}.} On the datasets testing a model's lexico-syntactic inference capabilities, the model 
performs below the majority class baseline. On the NER, EF, and Puns datasets its performs below the hypothesis-only baseline. 
We also notice that on three of the datasets 
(EF, Puns, and VN), 
the fixed hypothesis-only model outperforms the fixed \inferSent model.

These results might suggest that Multi-NLI does not evaluate whether sentence representations capture these distinct semantic phenomena. 
This is a bit surprising for some of the recast phenomena. 
We would expect Multi-NLI's fiction section (especially its humor subset) in the training set  
to contain some figurative language that might be similar to puns, and the travel guides (and possibly telephone conversations) to contain text related to sentiment. 

\noindent
\paragraph{Pre-training on DNC or Multi-NLI?} 
Initializing a model with parameters pre-trained on DNC or Multi-NLI often outperforms random 
initialization.\footnote{
Pre-training does not improve 
accuracies on NER or MV.}
Is it better to pre-train on DNC or Multi-NLI?
On five of the recast datasets, using a model pre-trained on DNC outperforms a model pre-trained on Multi-NLI. The results are flipped on the two datasets focused on downstream tasks (Sentiment and RE) and MV. However, the differences between pre-training on the DNC or Multi-NLI are small.
From this, it is unclear whether pre-training on DNC is better than Multi-NLI. 

\noindent
\paragraph{Size of Pre-trained DNC Data}
We randomly sample $10$K and $20$K examples from each datasets' training set to investigate what happens if we train our models on a subsample of each training set instead of the entire DNC. 
Although we noticed a slight decrease across each recast test set, 
the decrease was not significant. 
We leave this investigating
for a future thorough study. 

\section{Related Work}
\label{sec:related-work}
\vspace{-10pt}
\noindent
\paragraph{Exploring what linguistic phenomena neural models learn}
Many tests have been used to probe how well neural models learn different linguistic phenomena. \newcite{linzen2016assessing} use ``number agreement in English subject-verb dependencies'' to show that LSTMs 
learn about syntax-sensitive dependencies. 
In addition to syntax~\cite{shi-padhi-knight:2016:EMNLP2016}, researchers have used 
other labeling
tasks to investigate whether neural machine translation (NMT) models learn different 
linguistic phenomena~\cite{P17-1080,belinkov-EtAl:2017:I17-1,dalvi-EtAl:2017:I17-1,Marvin2018}. Recently, \newcite{poliakNAACL18} 
used recast NLI datasets 
to investigate semantics captured by  NMT encoders. 

\noindent
\paragraph{Targeted Tests for Natural Language Understanding}
We follow a long line of work focused on building datasets to test how well NLU systems perform distinct types of semantic reasoning. FraCaS uses a limited number of sentence-pairs to test 
whether systems understand semantic phenomena, e.g. generalized quantifiers, temporal references, and (nominal) anaphora~\cite{cooper1996using}. FraCas cannot be used to train neural models -- it includes just roughly $300$ high-quality instances
manually created by linguists.
\newcite{maccartney2009natural} created the \textit{FraCaS textual inference test suite} by automatically ``convert[ing] each FraCaS question into a declarative hypothesis.'' 
\cameraready{\newcite{levesque2012winograd}'s Winograd Schema Challenge forces a model to choose between two possible answers for a question based on a sentence describing an event.}

Recent 
benchmarks 
test whether
NLI models 
handle adjective-noun composition~\cite{P16-1204}, 
other types of composition~\cite{2018arXiv180204302D}, paraphrastic inference, anaphora resolution, and semantic proto-roles~\cite{white-EtAl:2017:I17-1}.
\cameraready{Concurrently, \newcite{conneau2018probe}'s benchmark can be used to probe whether sentence representations capture many linguistic properties. It includes syntactic and surface form tests but does not focus on as a wide range of semantic phenomena as in the DNC. 
\newcite{glockner-shwartz-goldberg:2018:Short} introduce a modified version of SNLI to test how well NLI models perform when requiring lexical and world knowledge.

\newcite{wang2018glue}'s GLUE dataset is intended to evaluate and potentially train a sentence representation to perform well across different NLP tasks. This continues an aspect of the initial RTE collection, designed to be representative of downstream tasks like QA, MT, and IR~\cite{dagan2010recognizing}.  While GLUE is therefore concerned with applied tasks, DNC, as well as \newcite{naik-EtAl:2018:C18-1}'s NLI \textit{stress} tests, is concerned with \emph{probing} the capabilities of NLU models to capture explicitly distinguished aspects of meaning.  
While one may conjecture that the latter is needed to be ``solved'' to eventually ``solve'' the former, it may be that these goals only partially overlap. Some NLP researchers might focus on probing for semantic phenomena in sentence representations while others may be more interested in developing single sentence representations that can help models perform well on a wide array of downstream tasks.}

\section{Conclusion}
\vspace{-5pt}
We described how we recast a wide range of semantic phenomena from many NLP datasets into labeled NLI sentence pairs. These examples serve as a diverse NLI framework that may help diagnose whether NLU models capture and perform distinct types of reasoning. Our experiments demonstrate how to use this framework as an NLU benchmark.
The DNC is actively growing as we continue recasting more datasets into labeled NLI examples. We encourage dataset creators to recast their datasets in NLI and invite them to add their recast datasets into the DNC. The collection, along with baselines and trained models
are available online at \website. 

\section*{Acknowledgements}
We thank Diyi Yang for help with the PunsOfTheDay dataset, 
the JSALT ``Sentence Representation'' team for insightful discussions, 
 and three
anonymous reviewers for feedback.
This work was supported by the JHU HLTCOE, DARPA LORELEI and AIDA, NSF-BCS (1748969/1749025), and NSF-GRFP (1232825).
The views and conclusions contained in this publication
are those of the authors and should not be
interpreted as representing official policies or endorsements
of DARPA or the U.S. Government.

\bibliography{references}
\bibliographystyle{acl_natbib}

\appendix

\title{Appendix for \textit{Collecting Diverse Natural Language Inference Problems for  Sentence Representation Evaluation}}

\section{More Recast NLI Examples}
\tabref{tab:app-nli_examples} includes examples from all of the recast NLI datasets. We include one \textsc{entailed} and one \textsc{not-entailed} example from each dataset that tests a distinct type of reasoning.

\label{app:recast-examples}
\begin{table*}[t!]
 \centering
 \small
 \renewcommand{\arraystretch}{1.35}
 \begin{tabular}{c|p{5.65cm}|p{5.65cm}}
     \toprule 
Semantic Phenomena & \multicolumn{1}{c|}{\cmark} & \multicolumn{1}{c}{\xmark} \\ \midrule 
& I would like to learn how & I'll not say anything \\ 
\multirow{-2}{*}{Event Factuality}
& The learning did not happen & The saying happened \\ 
\midrule 
 & Ms. Rice said the United States must work intensively &  Afghan officials are welcoming the Netherlands' decision \\ 
 \multirow{-3}{*}{Named Entity Recognition} 
 & Ms. is a person 's title & The Netherlands is an event \\ 
 \midrule 
 & The student met with the architect to view her blueprints for inspiration & The appraiser told the buyer that he had paid too much for the painting \\
 \multirow{-3}{*}{Gendered Anaphora}
 & The architect has blueprints & The appraiser had purchased a painting\\ 
 \midrule
 & Someone assumed that a particular thing happened & A particular person craved to do a particular thing \\ 
\multirow{-3}{*}{MegaVeridicality}
& That thing might or might not have happened & That person did that thing \\ 
\midrule
& The Romans destroyed the city & Andre presented the plaque \\
\multirow{-2}{*}{VerbNet} 
& The Romans caused the destroying & Andre was transferred \\
\midrule
& Molly wheeled Lisa to Rachel & Kyle bewildered Mark \\
\multirow{-2}{*}{VerbCorner} 
& Someone moved from their location & Someone or something changed physically \\
\midrule 
 & At least 100,000 Chinese live in Lhasa, outnumbering Tibetans two to one & Tropical storm Humberto is expected to reach the Texas coast tonight\\  
\multirow{-3}{*}{Relation Extraction}
& Tibetans live in Lhasa & Humberto hit Texas\\ 
\midrule
&  Jorden heard that my skiing skills are really going downhill & Caiden heard that fretting cares make grey hairs  \\ 
\multirow{-3}{*}{Puns} 
& Jorden heared a pun & Caiden heared a pun \\ 
\midrule
 & When asked about the product, Liam said, ``Don't waste your money'' & 
 When asked about the movie, Angel said, ``A bit predictable'' \\
 \multirow{-3}{*}{Sentiment Analysis}
 & Liam did not like the product & Angel liked the movie\\ 
 
\bottomrule 
 \end{tabular}
 
 \caption{Example sentence pairs for different semantic phenomena. The \cmark and \xmark columns respectively indicate that the context entails, or does not entail the hypothesis.
 Each cell's first and second line respectively represent a context and hypothesis.} 
 \label{tab:app-nli_examples}
 \end{table*}

\section{Recasting Semantic Phenomena}
\label{sec:app-recast}
Here we add secondary information about the original datasets and our recasting efforts.

\subsection{Event Factuality}
\label{app:ef}
We demonstrate how determining whether an event occurred can enable accurate inferences based on the event.
Consider the following sentences: 
\ex. 
\a. She walked a beagle
\label{ex:context-beagle}
\b. She walked a dog
\label{ex:hyp-beagle-dog}
\c. She walked a brown beagle
\label{ex:hyp-brown-beagle}

If the \textit{walking} occurred, \ref{ex:context-beagle} entails \ref{ex:hyp-beagle-dog} but not \ref{ex:hyp-brown-beagle}. If we negate the action in sentences  \ref{ex:context-beagle}, \ref{ex:hyp-beagle-dog}, and \ref{ex:hyp-brown-beagle} 
to respectively become: 
\ex. 
\a. She did not walk a beagle
\label{ex:premise-beagle-neg}
\b. She did not walk a dog
\label{ex:hyp-beagle-dog-neg}
\c. She did not walk a brown beagle
\label{ex:hyp-brown-beagle-neg}

The new hypothesis \ref{ex:hyp-brown-beagle-neg} is now entailed by the context \ref{ex:premise-beagle-neg} while \ref{ex:hyp-beagle-dog-neg} is not.

\verbose{\subsection{Named Entity Recognition}

\subsection{Gendered Anaphora Resolution}}

\subsection{Lexicosyntactic Inference}
\label{app:lexico}

\subsubsection{VerbCorner}
\label{app:vc}
When recasting VerbCorner, we use the following templates for hypotheses, assigning them as \textsc{entailed} and \textsc{not-entailed} based on the positive or negative answers to the annotation task questions about the context sentence.
\ex.
\a. Someone \{moved/did not move\} from their location
\label{vc:simonfreeze}
\b. Something touched another thing / Nothing touched anything else
\label{vc:explode}
\c. Someone or something \{applied/did not apply\} force onto something
\label{vc:equilibrium}
\d. Someone or something \{changed/did not change\} physically
\label{vc:entropy}
\d. Someone \{changed/did not change\} their thoughts, feelings, or beliefs
\label{vc:fickle}
\e. Something \{good/neutral/bad\} happened
\label{vc:goodworld}

\verbose{\subsection{Relation Extraction (KG Relations)}

\subsection{Subjectivity}}

\subsection{Figurative Language}
Puns in \newcite{D15-1284} were originally extracted from \url{punsoftheday.com}, 
and sentences without puns came from newswire and proverbs. The sentences are labeled as
containing a pun or not. Puns in \newcite{miller-hempelmann-gurevych:2017:SemEval} were sampled from prior pun detection datasets~\cite{miller2015automatic,miller2016towards} 
 and includes new examples generated from scratch for the shared task; the original labels denote whether the 
 sentences contain homographic, heterographic, or no pun at all. Here, we are only interested in whether a sentence contains a pun or not instead of discriminating between homographic and heterographic puns.
 
\subsection{Relation Extraction}
\label{app:kg}
Since hypotheses were automatically generated from Wikipedia infoboxes, many examples are noisy and ungrammatical. We presented hypotheses (independent of their corresponding contexts) to Mechanical Turk workers and asked them to label each sentence as containing no grammatical error, minor grammatical issues, or major grammatical issues. We removed the $2,056$ NLI examples with hypothesis containing major grammatical issues, resulting in $28,041$ labeled pairs. Interestingly, almost $70$\% of those examples where labeled between $1-4$, which we view as \textsc{not-entailed}. We release the ungrammatical NLI examples as supplementary data.

A second source of noise in the recast relation extraction dataset can be caused by disagreement amongst multiple annotators.
Examples in our training and development sets are annotated by a single annotator while we use $3$- to $5$-way redundancy to annotate the test examples. To guarantee high-quality test examples, we only include examples with 100\% inner-annotator agreement. Additionally, we remove the 16 examples labeled with $4$ from our \textsc{not-entailed} examples in this pruned test set since some of these examples are arguably entailments.  Consequently, the test set contains $761$ examples, out of the original $3,670$ test examples. Nevertheless, we separately release all $3,670$ test examples and include the original annotations as well, enabling others to consider other methods to collapse the multi-way annotations. 
 
\subsection{Sentiment}
\newcite{kotzias2015group} compiled examples from previous sources. The movie dataset came from \newcite{maas2011learning}, the Amazon product reviews were released by \newcite{mcauley2013hidden} add the restaurant reviews were sourced from the Yelp dataset challenge.\footnote{\url{http://www.yelp.com/dataset_challenge}}
 
\section{Experimental Details}
\label{app:sec-exp-details}
In all our experiments, we use pre-computed GloVe embeddings~\cite{pennington2014glove} and use the OOV vector for words that do not have a defined embedding.
We follow \newcite{conneau-EtAl:2017:EMNLP2017}'s procedure to train our models. During training, our models are optimized with SGD. Our initial learning rate is $0.1$ with a decay rate of $0.99$. Our models train for at most $20$ epochs and can optionally terminate early when the learning rate is less than $10^{-5}$. If the accuracy deceases on the development set in any epoch, the learning rate is divided by $5$.
As described in \newcite{hypoths-nli}, our hypothesis-only model feeds the hypotheses' encoded representation directly into the MLP.

\end{document}